\documentclass{amia}
\usepackage{graphicx}
\usepackage[labelfont=bf]{caption}
\usepackage[superscript,nomove]{cite}
\usepackage{color}
\usepackage{booktabs}       
\usepackage{wrapfig}
\usepackage{caption}
\usepackage{subcaption}
\usepackage{tabularx}
\usepackage{enumitem} 
\usepackage[colorinlistoftodos]{todonotes}
\usepackage{xcolor}

\begin{document}

\title{Is Deep Reinforcement Learning Ready for Practical Applications in Healthcare? A Sensitivity Analysis of Duel-DDQN for Hemodynamic Management in Sepsis Patients}

\author{MingYu Lu, MD, MBI$^{1}$, Zachary Shahn, PhD$^{2}$, Daby Sow, PhD$^{2}$
\\Finale Doshi-Velez, PhD$^3$, Li-wei H. Lehman PhD$^1$} 
\institutes{
    $^1$MIT, Cambridge, MA; $^2$IBM Research, Yorktown Heights, NY; $^3$ Harvard University, Cambridge, MA\\
}

\maketitle
\noindent{\bf Abstract}

\textit{
The potential of Reinforcement Learning (RL) has been demonstrated through successful applications to games such as Go and Atari. However, while it is straightforward to evaluate the performance of an RL algorithm in a game setting by simply using it to play the game, evaluation is a major challenge in clinical settings where it could be unsafe to follow RL policies in practice. Thus, understanding sensitivity of RL policies to the host of decisions made during implementation is an important step toward building the type of trust in RL required for eventual clinical uptake. In this work, we perform a sensitivity analysis on a state-of-the-art RL algorithm (Dueling Double Deep Q-Networks) applied to hemodynamic stabilization treatment strategies for septic patients in the ICU. We consider sensitivity of learned policies to input features, embedding model architecture, time discretization, reward function, and random seeds. We find that varying these settings can significantly impact learned policies, which suggests a need for caution when interpreting RL agent output.}

\section*{Introduction}
Artificial intelligence is changing the landscape of healthcare and biomedical research. 
Reinforcement Learning (RL) and Deep RL (DRL) in particular provide ways to directly help clinicians make better decisions via explicit treatment recommendations.  
Recent applications of DRL to clinical decision support include estimating strategies for sepsis management\cite{SARSA, AI_clinician,RLsepsis, DRL_for_sepsis,continouseState}, mechanical ventilation control \cite{RLonVentilation}, and HIV therapy selection\cite{HIV_treatment_selection}. However, the quality of these DRL-proposed strategies is hard to determine: the treatment strategies are typically learned from retrospective data without access to the unobservable counterfactual reflecting what would have happened had clinicians followed the DRL strategy. Unlike DRL strategies for Atari\cite{Atari} or other games \cite{DQN}, which can be evaluated by simply using them to play the game, testing a DRL healthcare strategy via a randomized trial can be prohibitively expensive and unethical. 

Thus, it is critical that we find ways to assess DRL-derived strategies prior to experimentation. One particular axis of assessment is robustness. DRL algorithms involve many choices, and if the output treatment policy is highly sensitive to some choice, that may imply either (a) getting that choice right is truly important or (b) we should be cautious about assigning credence to the output policy because a seemingly small and possibly unimportant change can have a large impact on results. In contrast, if the output policy is robust to analysis decisions then any errors are likely due to traditional sources of bias in observational studies (such as unobserved confounding), which can be more readily considered by subject matter experts assessing the credibility and actionability of DRL results.

In this paper, we explore the sensitivity of a particular DRL algorithm (duelling double Deep Q-networks, or Duel-DDQN\cite{double_qlearning,Dueling_DQN}, a state-of-the-art DRL method which has led to many success in scaling RL to complex sequential decision-making problems\cite{DQN})
to data preparation and modeling decisions in the context of hemodynamic management in septic patients. Septic patients require repeated fluid and/or vasopressor administration to maintain blood pressure, but appropriate fluid and vasopressor treatment strategies remain controversial \cite{ survivingSepsisCampaign,Inotropes_and_Vasopressors}. Past DRL applications by Komorowski,\cite{SARSA, AI_clinician} Raghu,\cite{continouseState,DRL_for_sepsis} and Peng \textit{et al}\cite{RLsepsis} make different implementation decisions while seeking to identify optimal fluid and vasopressor administration strategies in this setting (see additional discussion in Gottesman \textit{et al}\cite{evaluatingRLonHealth}).  However, these works do not perform systematic sensitivity analyses around their choices.

Starting with a baseline model similar to the works above, we perform sensitivity analyses along multiple axes, including: (1) inclusion of treatment history in the definition of a patient's state; (2) time bin durations; (3) definition of the reward; (4) embedding network architecture; and (5) simply setting different random initialization seeds. In all cases, we find that the Duel-DDQN is sensitive to algorithmic choices. In some cases, we have clear guidance: for example, making sensible decisions about a patient now requires knowing about their prior treatments. In other cases, we find high sensitivity with no clear physiological explanation; this suggests an area for caution and concern.

The paper is organized as follows. We first quickly review background and related work in DRL and sepsis and introduce some notation and terminology used throughout the paper. We then describe the components of a `baseline' DDQN implementation in detail. For select components that we examine in a sensitivity analysis, we discuss why different specifications could be reasonable and how we go about evaluating sensitivity to alternative specifications. Then we present results of our sensitivity analysis and conclude with a discussion highlighting several limitations and pitfalls to avoid when applying DRL in clinical settings.

\vspace{-0.5em}
\section*{Background and Related Work}\label{background}

\textbf{\textsl{\textit{Markov Decision Process(MDP) \& Q-Learning}}}
\vspace{-0.5em}

A Markov decision process (MDP) is used to model the patient environment and trajectories, which consists of\cite{Brief_surcey_DRL}
\begin{itemize}[noitemsep,topsep=0pt]
    \item A set of states $S$, plus a distribution of starting states $p(s_0)$.
    \item A set of actions $A$.
    \item Transition dynamics $T (s_{t+1}|s_{t},a_t)$ that map a state action pair at time t onto a distribution of states at time
    $t + 1$.
    \item An immediate/instantaneous reward function
    $ r_t = R(s_{t},a_{t},s_{t+1})$. 
    \item A discount factor 
     $\gamma \in [0, 1] $ , where lower values place more emphasis on immediate rewards.
\end{itemize} 
Every roll-out of a policy accumulates rewards from the environment, resulting in the return $ R = \sum^{T-1}_{t=0}\gamma^tr_{t+1}$. The goal of RL is to find an optimal policy $\pi^{*}$, which achieves the maximum expected return from all the states. $ \pi^{*} = \arg\max_{\pi}E[R|\pi]$. To find $\pi^{*}$, one of the reinforcement learning algorithm is $Q$-learning. The basic idea to evaluate the policy is to use temporal difference (TD) learning over the policy iteration to minimize the TD error\cite{Qlearning,Brief_surcey_DRL}. 
$Q^{\pi}(s,a) \leftarrow Q^{\pi} + \alpha(r +\gamma +  Q^{\pi}(s',a') - Q^{\pi}(s,a) )$

More formally, Q learning aims to approximate the optimal action-value function given the observed state $s$ and the action $a$ at time $t$\cite{Qlearning, Brief_surcey_DRL}. The future reward $r_{t}$ is discounted at every time step $t$ by a constant factor.
$ Q^{*}(s,a) = E[r_t + \gamma*r_{t+1} + \gamma^2* r_{t+2} +... | s_t, a_t= a,\pi]$

\textbf{\textsl{\textit{Dueling Double Deep Q-Learning (Dueling DDQN) with Prioritized Experience Relay (PER)}}}
\vspace{-0.5em}

To approximate the optimal action-value function, we can use a deep Q-network: $Q(s,a;\theta)$ with parameter $\theta$. To estimate this network, we optimize the following sequence of loss functions at iteration $i$: $L_{i}(\theta_{i}) = E_{s,a,r,s'}[(y_{i}^{DQN} - Q(s, a; \theta_{i}))^2] $;$y_{i}^{DQN} = r + \gamma maxQ(s', a'; \theta'),$ updating parameters by gradient descent such that $\nabla\theta_{i}L_{i}(\theta_{i}) = E_{s,a,r,s'}[y_{i}^{DQN} - Q(s, a; \theta_{i}) \nabla\theta_{i}Q(s, a; \theta_{i})].$

Dueling Double Deep Q-learning\cite{Dueling_DQN} is a particular state-of-the-art deep Q-learning algorithm consisting of separate `dueling' architectures that decouple the $value$ and $advantage$ streams in deep Q-networks\cite{Dueling_DQN} to determine the value of the next state\cite{double_qlearning}. Prioritized experience replay\cite{double_qlearning, prior_experience_relay}, i.e. sampling mini-batches of experience that have high expected impact on learning, further improves efficiency.

\textbf{\textsl{\textit{Learning Sepsis Management with DRL}}}  
\vspace{-0.5em}

Sepsis is a life-threatening organ dysfunction disease caused by dysregulated host response to infections \cite{survivingSepsisCampaign}. How to maintain septic patients' hemodynamic stability via administration of intravenous fluid (IV fluid) and vasopressors is a key research and clinical challenge.\cite{survivingSepsisCampaign,Inotropes_and_Vasopressors} A number of DRL studies have been carried out to address this issue in the past few years. Raghu \textit{et al}\cite{DRL_for_sepsis,continouseState} applied a Dueling Q Network with sparse autoencoder. Peng \textit{et al}\cite{RLsepsis} further presented a mixture-of-experts framework, combining a kernel and a Dueling DDQN with PER to personalize sepsis treatment. It is worth noting that these studies were conducted based on different reward settings. Raghu \textit{et al}\cite{DRL_for_sepsis,continouseState} used hospital and 90-day mortality as a sparse reward issued at the end of patients' trajectories, and subsequent analysis used short-term rewards such as SOFA score in combination with lactate levels \cite{DRL_for_sepsis}, while Peng \textit{et al}\cite{RLsepsis} used changes in probability of mortality. 


\renewcommand{\thefootnote}{$*$}

\textbf{\textsl{\textit{Data Description \& Cohort} \label{cohort}}}
Data for our cohort were obtained from the Medical Information Mart for Intensive Care (MIMIC-III v1.4)\cite{MIMIC} database. The data set contained all MIMIC-III patients meeting Sepsis-3 criteria \cite{Sepsis-3} from years 2008-2012.\footnote{ICU admissions between 2008-2012 were recorded using the MetaVision system with higher resolution treatment information.} It comprised 7,956  patients with 649,661 clinical event entries. In this analysis, we extracted and collected static features (e.g. demographic), past treatment history and a summary of hourly observation (mean, maximum, and minimum within an hour) of all laboratory values within patients' first 72-hour ICU stay. For intravenous fluids, we extracted commonly used IV fluids for resuscitation (colloids and crystalloids). The detailed features are shown in Table 1.  All measured values were standardized, and we carried forward covariate values from the most recent measurement. The data set was split using 80\% for training and validation and 20\% for testing.

\vspace{0.5em}

\begin{table}[H]
\captionsetup{font=small, skip=0.5em}
\caption{Details of attributes including vital signs, laboratory values, and treatment history}
    \centering
    \scalebox{0.8}{
    \begin{tabularx}{1.2\textwidth}{lX}
    \toprule
      Demographic   & Age, Weight, Height, Ethnicity \\
      \midrule
      Vital Sign\textbackslash Laboratory & GCS, Heart rate, Temperature, Respiratory Rate, Diastolic Blood Pressure, Systolic Blood Pressure, Mean Arterial Blood Pressure, Potassium, Sodium, Chloride, Magnesium, Calcium, Anion gap, Hemoglobin, Hematorcit, WBC, Platelets, Bands, PTT, PT, INR, Arterial pH, $SpO_{2}$, $FiO_{2}$, $PaO_{2}$, $Total CO_{2}$, $pCO_{2}$, Arterial Base excess, Bicarbonate, Arterial Lactate, SOFA score, Glucose, Creatinine, BUN, Total Bilirubin, Indirect Bilirubin, Direct Bilirubin, AST, ALT, Total Protein, Troponin, CRP, Elixhauser Score, Albumin \\ 
      \midrule
      Treatment & Vasopressor \& IV Fluid \\
    \bottomrule
    \end{tabularx}}
\end{table}

\vspace{-1.5em}
\section*{Methods: Baseline Implementation}
We first describe our baseline implementation of a Dueling DDQN with PER to learn an optimal resuscitation strategy. This implementation combines elements from several published RL applications to sepsis treatment\cite{continouseState,RLsepsis}, with slight modifications. In describing the baseline, we also illustrate the many components involved in specification of a Dueling DDQN analysis. In our sensitivity analysis, we will systematically vary these components from our baseline model.

%



\begin{wrapfigure}{r}{0.4\textwidth}
    \centering
    \vspace{-1.4 em}
    \includegraphics[scale=0.3]{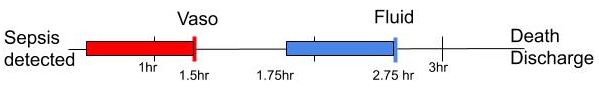}
    \captionsetup{font=footnotesize,skip=0.5em}
    \caption{Time discretization}  
    \label{fig:time_bin}
\end{wrapfigure}

\textbf{\textsl{\textit{Time Discretization }}}
We divided patient data into one-hour bins. To avoid inappropriately adjusting for covariate values that were measured after treatment actions were taken \cite{time_varying_instructional_treatments}, we performed a time-rebinning procedure. If a treatment action occurred within an existing time bin, covariate measurements made after the treatment action in that bin were moved to the following bin and the time of the treatment action became the new endpoint of the time bin. Figure \ref{fig:time_bin} illustrates this process with the blue bar covering 1.75 hr to 2.75 hr defining the time period for which covariate measurements contribute to the bin ending at 2.75 hours. (Time rebinning is rare in RL literature but necessary to avoid adjusting for post-treatment variables.)


\textbf{\textsl{\textit{Compressing Patient Histories }}} We follow Peng \textit{et al}. \cite{RLsepsis} in encoding patient states recurrently using an LSTM autoencoder representing the cumulative history for each patient. LSTMs can summarize sequential data through an encoder structure into a fixed-length vector and then reconstruct into its original sequential form through the decoder structure\cite{LSTM}. The summarized information can be used to represent time series features\cite{recurrent_auto_encoder_for_multidimentional_time_series_representation}. LSTM-RNN models can prevent a vanishing or exploding gradient and are commonly used to capture long-term sequence structures\cite{Empirical_Evaluation_of_Gated_Recurrent_Neural_Networks_on_Sequence_Modeling}.

\textbf{\textsl{\textit{Action definition and Treatment Discretization  }}} Following  Raghu \textit{et al}. \cite{continouseState} and Peng \textit{et al}.\cite{RLsepsis}, we focus on intravenous fluids and vasopressors as the actions of the MDP. We computed the hourly rate of treatment as the action and sum the rate when there are overlapped treatment events of the same type. The hourly rate of each treatment is divided into 5 bins defined by quartiles under current physician practice. Accordingly, a 5 by 5 action space is defined for the medical intervention \cite{continouseState}. An action of (0,0) means no treatment is given; whereas, an index of (4, 4) represents top quartile dosages of both fluids and vasopressors. 

\textbf{\textsl{\textit{Reward formulation  }}} We follow Peng \textit{et al.}\cite{RLsepsis} in defining the reward at time $t$ as the change in negative log-odds of 30 day mortality between $t$ and $t+1$ according to a predictive model for 30 day mortality. The probability of mortality was estimated with a 2-layer neural network with 50 and 30 hidden units with L1 regularization given the recurrent embedding of the compressed history at the corresponding time. Let $f(o)$ be the probability of mortality given observations through the current time point $o$ and $f(o')$ be the probability of mortality given observations through the next time step. Then we define the reward
\vspace{-1em}
\begin{equation}
r(o,a,o') = -log\frac{f(o')}{1-f(o')} + log\frac{f(o)}{1-f(o)}.
\end{equation}

\vspace{-1em}


\textbf{\textsl{\textit{Dueling DDQN Architecture  }}}  Following Raghu \textit{et al} \cite{continouseState}, our final Duelling Double-Deep Q Network (Dueling DDQN)\cite{Dueling_DQN,double_qlearning } with PER architecture has two hidden layers of size 128, using batch normalization after each, Leaky-ReLU activation functions, a split into equally sized advantage and value streams, and a projection onto the action-space by combining these two streams. The Duel-DDQN architecture divides the value function $V$ into the value of the patient’s underlying physiological condition, called the $Value$ stream, and the value of the treatment given, called the $Advantage$ stream.

\section*{Methods: Sensitivity analysis}\label{sensitivity}

In this section, we describe the ways in which we altered the baseline Dueling DDQN implementation described in the previous section in our sensitivity analysis. For each component that we varied, we specify the alternatives we considered, explain why the choice could be important, and also explain why each alternative might be considered reasonable. We emphasize that the aim of this  sensitivity analysis is not to determine which choices are best, as there is no ground truth available to make such a determination. Rather, it is to understand the robustness of the learned treatment policy with respect to a priori reasonable-seeming alternatives. In particular, we explore the effects on learned policies of: including treatment history in the state definition; varying the time bin size; varying the reward specification; specifying different recurrent embedding models; and setting different random seeds.

\textbf{\textsl{\textit{Including Past Treatment History}}}
\vspace{-0.5em}

One decision is whether to include the history of past treatments in the representation of patient state. Several prior works \cite{RLsepsis, DRL_for_sepsis,RLonVentilation} did not do so, but given the Markov assumption on which DQNs rest, this amounts to assuming that past treatments cannot impact future outcomes through pathways that do not run through measured covariates included in the state. This assumption will usually be false. For example, vasopressors have potentially serious long term cardiovascular side effects\cite{Inotropes_and_Vasopressors}, but the added risk after administering more vasopressors would not be captured in short term changes in measured patient covariates. Thus, in our sensitivity analysis we compare learned policies from DQN implementations with state summaries that include and exclude
treatment history. 
In this analysis, we consider the cumulative dosage of all previous treatment (IV fluid and vasopressor) until $t-1$ as a proxy for treatment history at $t$. 

\textbf{\textsl{\textit{Duration of Time Bins}}}
\vspace{-0.5em}

When applying a discrete time RL algorithm to data with actions taken and measurements recorded in continuous time, an implementation decision that inevitably arises is how to bin time into discrete chunks in which to define patient states $S_t$.   
With infinite data, shorter time bins would be superior for two reasons. First, the state at each time step reflects the patient's condition closer to when the treatment action at that time step was decided. This improves confounding adjustment, since states are more reflective of information that actually influenced treatment decisions. Second, more time steps allow for more flexible and dynamic learned strategies that are more responsive to changes in patient state. However, with finite data, the capacity to learn more flexible strategies with shorter time bins can be detrimental, leading to instability of the estimated optimal policy. 

In the case of sepsis, past work has used 4 hour time bins\cite{continouseState}. Treatment decisions in this clinical context are made on a finer time scale than 4 hours, which is why we defined 1 hour time bins in our baseline model. But stability is also a major concern in this dataset, so either choice is defensible. Hence, we compared the learned policies of Dueling DDQNs fit to 1 hour and 4 hour time binned data sets in our sensitivity analysis.


\textbf{\textsl{\textit{Horizon of Rewards}}}
\vspace{-0.5em}

A key decision is specifying the reward function. Ideally, the reward function would summarize the entire long term utility of the stay, as this is what we really seek to optimize. However, for reasons of practicality, researchers often choose short term rewards measured at each time-step. When rewards are short term, there is more `signal' in that it is easier to estimate associations between rewards and actions. 
If a RL algorithm is truely robust, the learned policies would be broadly similar whether we choose to optimize our true reward of interest or a shorter term proxy. To investigate the impact of using long term against short term reward, we also compared reward functions that were weighted mixtures of long and medium term information about outcomes for a range of weights. 

We define a utility function reward $U$ as follows. Let \textbf{$M$} be the worst possible SOFA score. 
Let \textbf{$Y$} be observed SOFA at the end of the stay.
Let $S = 1$ if the patient survived more than 1 year after admission, 0 otherwise.
Let $H$ be hours survived after admission.
Let $C$ be a constant that controls relative weight assigned to SOFA score at the end of stay and survival. We define $r'(C)$ as 
\vspace{-1em}
\begin{equation}
      \textbf{if } H \geq 24*365 
      \textbf{ then } U = log(1 + \frac{M - Y}{C}),
      \textbf{ else }  U = log(\frac{H}{24*365} + 1) 
\end{equation}
For large values of $C$, survival time is all that matters. For low values of C, patient state at the end of the stay matters a lot for patients who survive more than 1 year. For all values of $C$, rewards are medium to long term (since they are based on patient state at the end of the stay or later as opposed to the following time bin), but differing $C$ levels reflect different subjective prioritization of patient health outcomes.
We compare learned DQN policies using short term reward (1) with long term rewards (2) for varying values of $C$.


\textbf{\textsl{\textit{Choice of Embedding Model}}}
\vspace{-0.5em}

Another question is how to summarize patient history.
In DQNs, it is important for the information contained in the state $S_t$ at each time $t$ to satisfy several conditions. First, $S_t$ should satisfy the ``sequential exchangeability'' assumption\cite{Robust_estimation}, which would be satisfied if $S_t$ contains all relevant information about variables influencing treatment decisions at time $t$ and associated with future rewards, i.e. $S_t$ should contain sufficient information to adjust for confounding. If sequential exchangeability fails, then estimates of the impact of actions on future rewards will be biased, and therefore the estimate of the optimal treatment strategy will be biased as well.

To learn an optimal treatment strategy, it is also important that $S_t$ contain relevant information about variables that are \emph{effect modifiers}. An effect modifier is a variable with the property that the conditional average effect of an action on future rewards varies with the variable's value. Good treatment rules assign treatment based on the value of effect modifiers. (Effect modifiers may or may not be confounders, which are necessary to include in the model to avoid bias but may not be good inputs to treatment decision rules.) 


Finally, DQNs make a very strong Markov assumption on states\cite{Qlearning, Atari}. $S_t$ must be defined to be sufficiently rich that this Markov assumption can approximately hold. Thus, to allow for realistic long term temporal dependencies, states at each time should be rich summaries of patient history. Without a priori knowledge of exactly which aspects of patient history to retain (to adjust for confounding, model effect modification, and satisfy the Markov assumption), a reasonable strategy is to define patient states as embeddings generated by a RNN \cite{Reinforcement Learning with Long Short-Term Memory,recurrent_auto_encoder_for_multidimentional_time_series_representation}. 

However, RNN embeddings are not optimized to retain the types of information specifically required to be contained in DQN states. Different choices of black box embedding method may generate states that satisfy the DQN requirements to varying degrees and produce different learned policies, with no principled way to choose between them. 

In our sensitivity analysis, we compare two common RNN embedding models--long short-term memory (LSTM)\cite{LSTM} and gated recurrent unit (GRU)\cite{GRU}. The architectures have been shown to perform comparably across a range of tasks \cite{Empirical_Evaluation_of_Gated_Recurrent_Neural_Networks_on_Sequence_Modeling}. Each consists of two hidden layers of 128 hidden units and is trained with mini-batches of 128 and the Adam optimizer for 200 epochs or until convergence. We have no reason a priori to believe that either option would produce more suitable embeddings than the other, and the point of comparing them is to determine whether the decision is important.

\textbf{\textsl{\textit{Random restarts}}}
\vspace{-0.5em}

Finally, solving a DDQN is a non-convex optimization, and thus random restarts are frequently used to find a good local optimum.  
As Henderson \textit{et al}\cite{Deep_Reinforcement_Learning_that_Matters} reports, a wide range of results can be obtained from the same deep RL algorithm depending on the random initialization of the network's weights initialization. To observe the impact of random weight initialization in our dataset, we fit our baseline model repeatedly using different seeds.  While one would generally simply take the best of the random restarts as the solution, high variation across random restarts might mean that reproducing a result will be more challenging as the problem has many diverse local optima.

\section*{Methods: Evaluation Metrics \& Experimental Settings}

\begin{wrapfigure}{r}{0.22\textwidth}
    \centering
    \vspace{-2.6em}
    \includegraphics[scale=0.2]{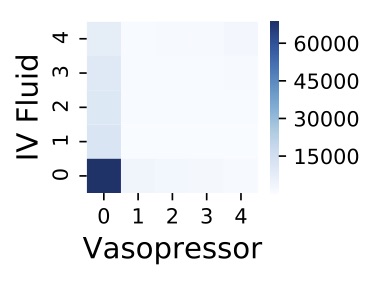}
    \captionsetup{font=footnotesize, skip=0em}
    \caption{Action distribution}
    \label{metrics}
\end{wrapfigure}

\textbf{\textsl{\textit{Metrics}}}
Previous works have used off-policy estimators, such as Weighted Doubly Robust (WDR\cite{DRLPolicyEvaluation}) to estimate the quality of a proposed policy. However, these estimators can have high variance as well as bias. Instead, we compare policies based on the distribution of actions they recommended. If these distributions are very different from each other or from clinicians (whom we know act reasonably) then that may be a cause for skepticism.  

Specifically, for each time point in the patient history in the test set, we compute the Dueling DDQN's recommended action. The action distribution for a policy is simply the frequency with which each of the 25 combinations of vasopressor and fluid doses are recommended across all person-times in the test set. We use heat maps like Figure \ref{metrics} to display action distributions, where the y axis represents the IV fluid dosage (quartile), the x axis represents the vasopressor dosage (quartile), and the density of the color represents the frequency with which the treatment action is applied across all person-times in the test set. These action distributions are aggregates in that we are summing over all time points and patients. 
Comparison 
of aggregate action distributions between different experimental settings can provide insights into the ways in which policies differ. \color{black}


 

\textbf{\textsl{\textit{Parameters and Optimization}}}
We train the  Dueling DDQN for 100,000 steps (except the reward horizon experiment, where we perform early stopping at 15,000 steps to prevent over-fitting) with a batch size of 30. We conducted 5 restarts for every experimental setting (except the random restart experiment, where we look at variation across individual restarts). Following Peng \textit{et al}\cite{RLsepsis}, of the policies resulting from the 5 restarts we choose the one with highest value as estimated by a weighted doubly robust off policy evaluation method \cite{DRLPolicyEvaluation}. For models trained with long term rewards, $r'(C)$, where the WDR estimator is unfeasible, we selected a policy from the 5 restarts based on the Q-value. 
\vspace{0.2em}
\begin{table}[h!]
    \captionsetup{font=small,skip= 0em}
    \caption{ Summary of the variance across different experimental settings}
   \centering
    \scalebox{0.8}{
        \begin{tabularx}{1.2\textwidth}{llllX}
            \toprule
            {}  & Timing & Encoder & Treatment history&  Reward  \\
            \midrule
            Baseline & 1 hr & LSTM (full history) & Yes& Short term: (1) Immediate change in prognosis \\
            Alternatives &  1hr, 4hr & LSTM\textbackslash GRU (full history) & Yes, No &  
            Short term: (1) Immediate change in prognosis; Long term: (2) Combinations of SOFA at end of stay and survival time \\
            Raghu \textit{et al}\cite{continouseState} & 4 hr & Sparse Autoencoder (current obs. only)& No & Long term: In-hospital mortality \\
            Peng \textit{et al}\cite{RLsepsis} & 4 hr & LSTM (full history) & No & Short term*: (1) Immediate change in prognosis \\
            \bottomrule 
            \multicolumn{5}{l}{ *  In Peng \textit{et al}'s reward, prognosis was estimated conditional on current observations only. In the baseline implementation's reward,} \\
            \multicolumn{5}{l}{ prognosis was estimated conditional on full patient history.}  \\
        \end{tabularx}
        }
\end{table}


\vspace{-1.4em}
\section*{Results}\label{results}

We altered aspects of the baseline  Dueling DDQN implementation described in the previous section and compared the resulting learned policies according to their action distributions. 
In the following, we abbreviate Dueling DDQN trained with embedding to DQN-embedding. For, example the Dueling DDQN trained with LSTM and 1 hourly binned data is called DQN-LSTM-1hr.



\textbf{\textsl{\textit{Treatment History: Excluding treatment history leads to aggressive treatment policies}}}
\vspace{-0.5em}

\begin{figure}[!h]
\centering
\includegraphics[scale=0.2]{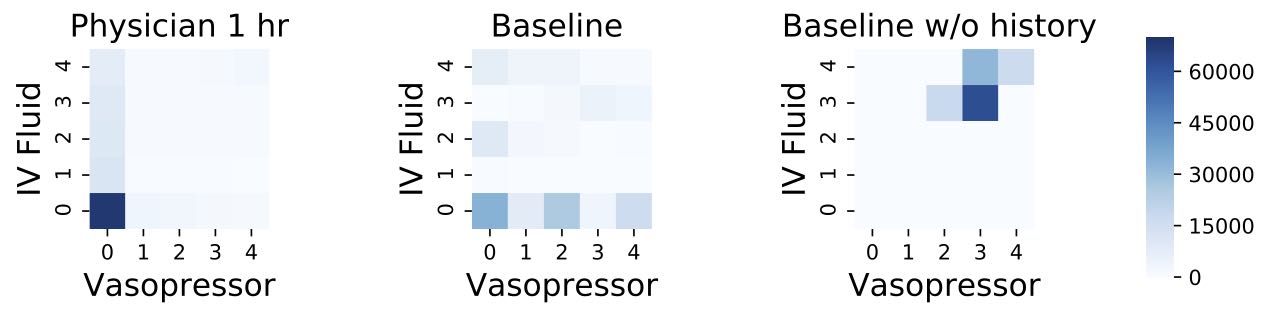}
\captionsetup{font=small, skip = 0em }
\caption{The action distribution of  Dueling DDQNs trained with/without past treatment history information. Note that the agent trained without treatment history aggressively prescribes high (3rd quartile, 4th quartile) dosage of vasopressor and IV fluid. Low dosage treatment are rarely administered to patients. }
\label{treatment_history}
\end{figure}

Here, we compare treatment strategies output by Dueling DDQNs that do and do not include treatment history in patient state representations (Figure \ref{treatment_history}). The DQN-LSTM-1hr trained without treatment history recommends nonzero vasopressor doses at all time points. It frequently recommends high doses of each treatment compared to physicians and an agent trained with treatment history included in the state definition. Excluding past treatment information increases the frequency of average recommended dosage of vasopressor and fluid by 1.6 - 1.8 times and 1.7 - 3.1 times, respectively in the test set. We hypothesize an explanation for this behavior in the Discussion section. 
\vspace{-0.7em}

\textbf{\textsl{\textit{Time bin durations: Longer time bins result in more aggressive policies.}}}
\vspace{-0.5em}

\begin{figure}[h!]
    \centering
    \vspace{-1em}
    \includegraphics[scale=0.2]{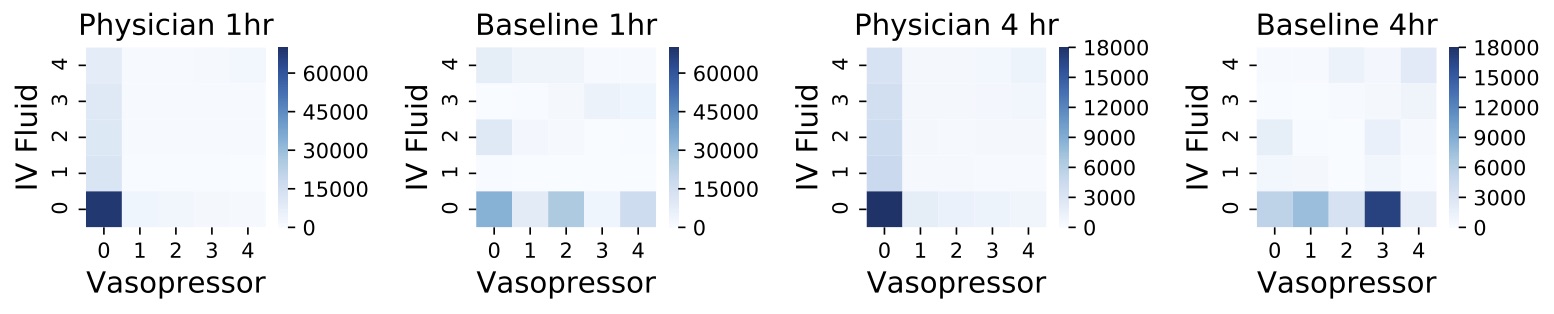}
     \captionsetup{font=small, skip=0em}
     \caption{Comparison of action distribution across DQN-LSTM 4-hour and DQN-LSTM 1-hour time. (Note the same color density does not represent the same count in 4 hour bins and 1 hour bins). The 4 hour bins lead to much more frequent recommendations of high vasopressor doses by the DQN-LSTM, while the physician's policy remains conservative.}
    \label{fig:diff_timebin_distribution}
\end{figure}

Figure \ref{fig:diff_timebin_distribution} illustrates that while different ways of segmenting time do not affect the clinician action distributions (by definition), they have a large effect on the  Dueling DDQN action distributions. The DQN-LSTM's 4 hour bins increased the frequency of nonzero vasopressor doses by 40\% and decreased
the overall usage of IV fluid only, a less aggressive action, by 34\% compared to baseline settings. 

\textbf{\textsl{\textit{Rewards: Long-term objectives lead to more aggressive and less stable policies}}}

\begin{figure}[!h]
    \centering
    \vspace{-0.8em}
         \includegraphics[scale=0.2]{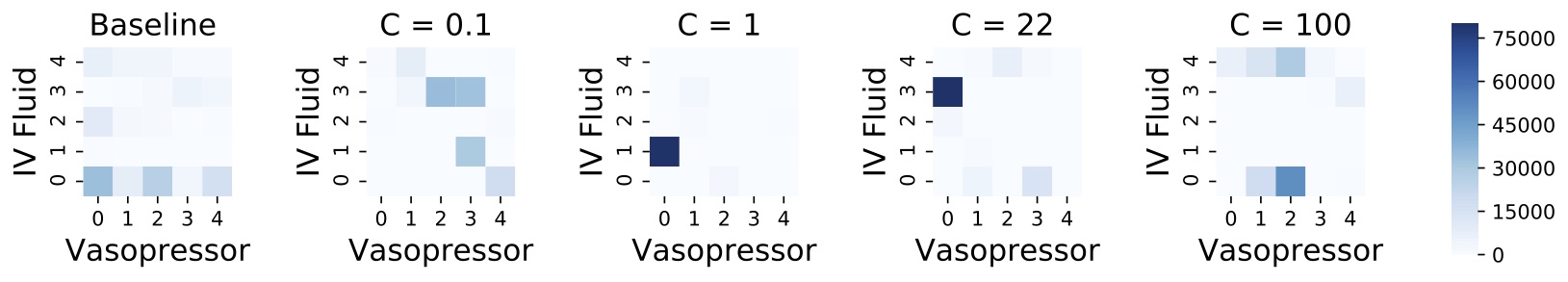}
         \captionsetup{font=small, skip= 0em}
        \caption{Comparison of Duel-DDQN trained with short term \& long/intermediate reward across varying Cs.}
\label{fig:reward-results}
\end{figure}

\vspace{-0.5em}
In figure~\ref{fig:reward-results}, we see that longer term objectives resulted in more aggressive policies---specifically in more frequent high fluid doses than our short term baseline reward for all values of $C$ (for C = 100, it is difficult to see visually in the heat map, but the agent administered the maximum fluid dosage 40\% of the time). Policies also vary considerably across level of emphasis on medium versus long term outcomes determined by values of $C$. We noticed higher variation across random restarts in the long term reward settings than the short term baseline settings (see Figure 8). This could indicate that optimization is more challenging and unstable for long term rewards. 

\textbf{\textsl{\textit{Embedding model: High sensitivity to architecture}}}

\begin{figure}[h!]
    \centering
    \includegraphics[scale=0.22]{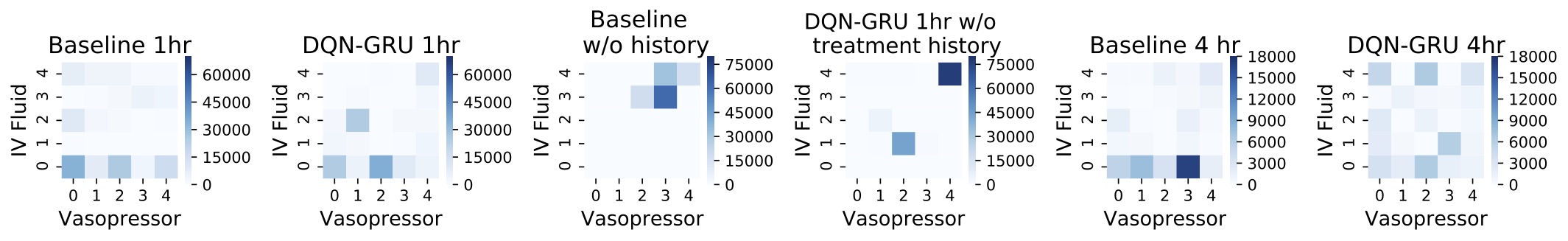}
    \captionsetup{skip= 0em, font=small}
    \caption{ \textsl{\textit{\textbf{GRU implementation}}}: Comparing to the baseline (LSTM), the most observed treatment by DQN-GRU 1hr was to deliver a medium dosage vasopressors without IV fluid;
    \textsl{\textit{\textbf{Exclusion of Treatment history}}}: DQN-GRU administers maximum dosage of both vasopressor and IV fluid most of the time; \textsl{\textit{\textbf{Time bin duration}}} : 4-hour bins increased the overall usage of IV fluid by 20\% and more than doubled increased maximum dosage of IV fluid.  }
    \label{different_embeddings}
\end{figure}

\vspace{-0.5em}
Results comparing LSTM and GRU embeddings can be found in Figure \ref{different_embeddings}. We can observe that both our baseline (LSTM) implementation and the GRU implementation recommended nonzero doses of vasopressors significantly more frequently than physicians. However, the GRU implementation was more aggressive, recommending nonzero fluid doses significantly more often than both physicians and the DQN-LSTM.

 The choice of embedding architecture also interacts with other analysis settings, whose effects differ depending on embedding architecture. We illustrate interactions with treatment history and time segmentation.

\emph{Exclusion of Treatment History} Excluding prior treatment history has an even more extreme effect when embeddings use the DQN-GRU architecture, with maximum dosage of both treatments being delivered most of the time.

\emph{Different time segmentation} 4 hour time bins led to more frequent high vasopressor doses in the baseline LSTM implementation, but more frequent high fluid doses in the GRU implementation. 

There is no way to apply clinical, physiological, or statistical knowledge to reason about which embedding architecture is more appropriate as  they are conceptually quite similar. Thus, the variation stemming from the choice of embedding is a source of concern. 

\vspace{-1em}
\vspace{0.5em}

\vspace{-0.5em}
\textbf{\textsl{\textit{Random Restarts: DRL policies have many local optima}}}
\vspace{-0.5em}

\begin{figure}[h]
    \begin{subfigure}[t]{0.56\textwidth}
    \centering
    \includegraphics[scale=0.19]{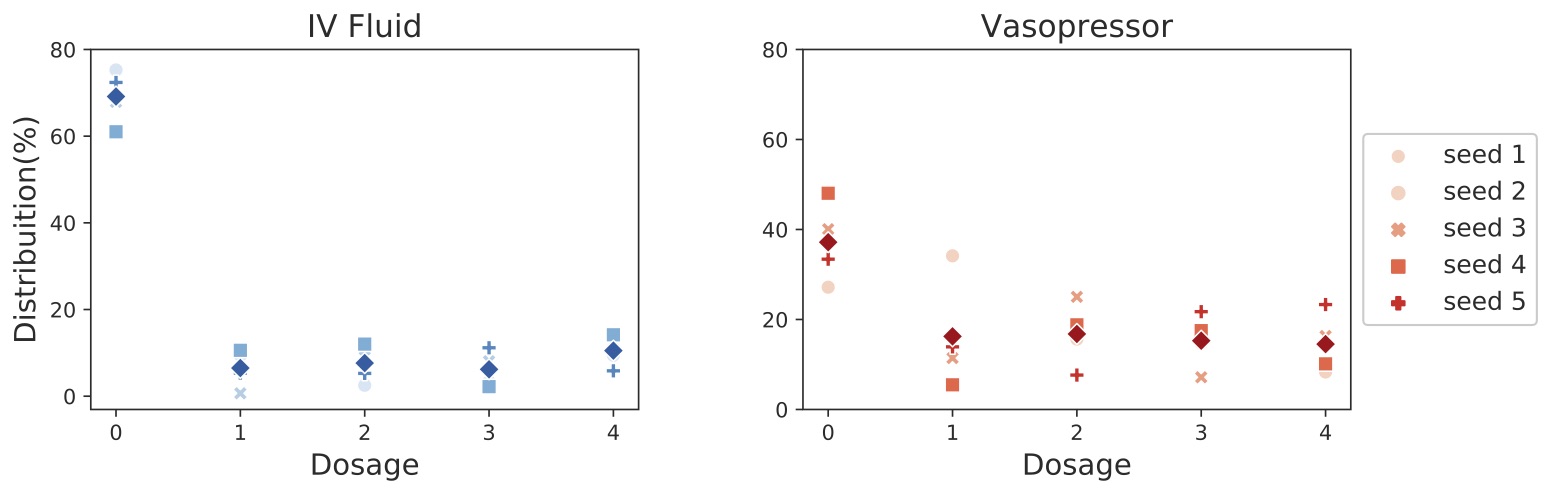}
    \captionsetup{skip=-0.3em}
    \caption{ }
    \label{random_restart_action}
    \end{subfigure}
    \quad
    \begin{subfigure}[t]{0.4\textwidth}
    \centering
    \includegraphics[scale=0.16]{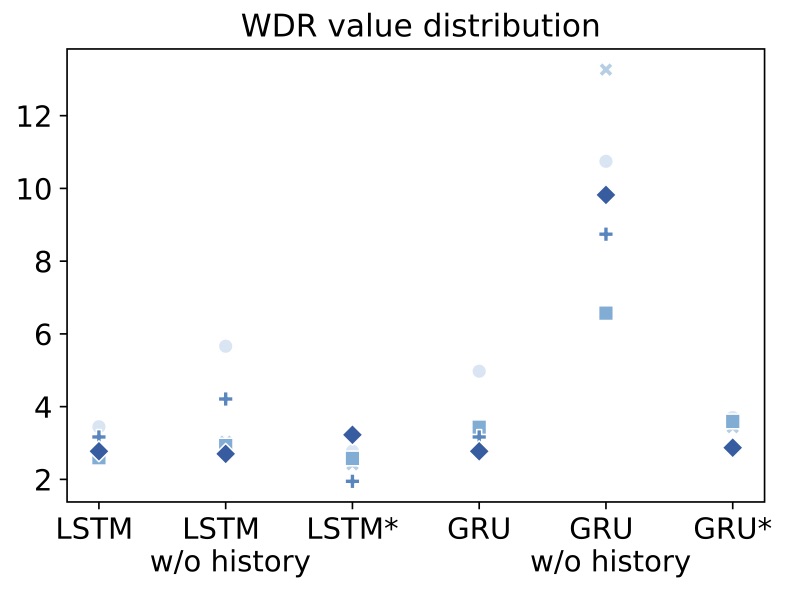}
    \captionsetup{skip=-0.3em}
    \caption{ }
    \label{random_restart_WDR}
    \end{subfigure}
    \captionsetup{belowskip= -1em,justification = raggedright, singlelinecheck = false, font=small}
    \caption{ (a) Treatment distribution across random restarts in baseline. While variance of IV fluid is small, distribution of vasopressor varies across different seeds. (b) Comparision of values of WDR estimator across random seeds in each settings (* represents 4 hour time bins duration). In the setting of exclusion of treatment history, agents are highly sensitive to the seeds.}
\end{figure}

Our final sensitivity analysis looked at variation across restarts of the algorithm, which assesses sensitivity to where the algorithm was initialized.  While the  
substantial differences in action distribution 
between the Dueling DDQN and physician policies remained constant across seeds in our baseline model, there was still much variation in the resulting action distributions, especially for vasopressors (Figure \ref{random_restart_action}). In Figure \ref{random_restart_WDR}, we see that despite these differences, the estimated values of these policies are similar; this demonstrates that the variation is not because the optimization sometimes landed in a poor optimum, but because there are many optima with similar estimated quality that lead to qualitatively different policies. This is another cause for concern, as it suggests that the agent has no way of telling if any of these very different policies are better than the others.



\vspace{-1em}

\begin{figure}[h]
    \begin{subfigure}[t]{0.5\textwidth}
    \centering
    \includegraphics[scale=0.16]{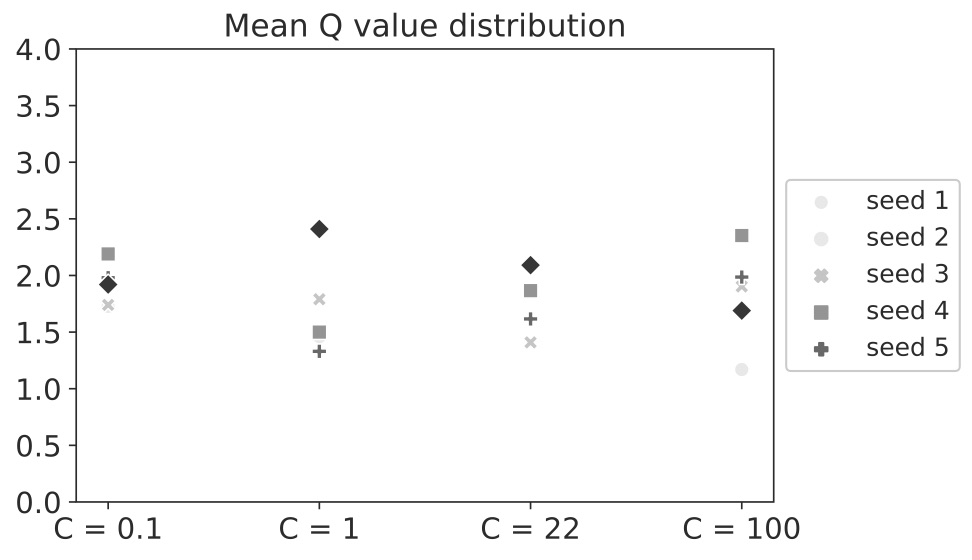}
    \captionsetup{skip=0em}
    \caption{ }
    \label{q_values}
    \end{subfigure}
    \quad
    \begin{subfigure}[t]{0.4\textwidth}
    \centering
    \includegraphics[scale=0.15]{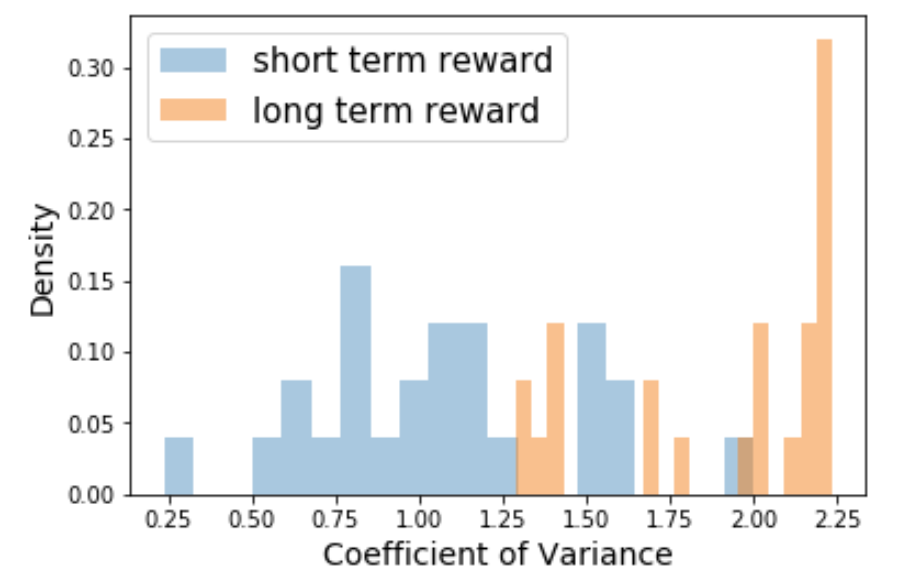}
    \captionsetup{skip=0em}
    \caption{ }
    \label{variance_across_restart}
    \end{subfigure}
    \captionsetup{ justification = raggedright, singlelinecheck = false, font=small}
    \caption{ (a) Q values distribution across different seeds (b)  
    Comparison between long term and short term (baseline) objectives. Note that long term reward implementation demonstrates higher variance compared to baseline. X-axis labels coefficient of variance or relative standard deviation, $ c_v = \frac{\sigma}{\mu}$ where $\sigma$ is the standard deviation and $\mu$ is the mean of action distribution across random restarts of each of the 25 discretized actions (as in Figure\ref{metrics}.)  
    }
\end{figure}

We also found that policies with medium and long-term objective policies were much more sensitive to random seed. Figure \ref{variance_across_restart} depicts the distribution over the action space of the variance across 5 random restarts of the frequency with which that action was recommended. Despite the similar estimated Q values across random seeds (Figure \ref{q_values}), the variances were much greater for the implementations using long term rewards. 
The presence of many local optima increases variance and makes it more challenging to differentiate policies in implementations with long term rewards.


   
\vspace{-0.5em}
\vspace{0.5em}
\begin{wrapfigure}{r}{0.45\textwidth}
    \centering
    \includegraphics[scale=0.19]{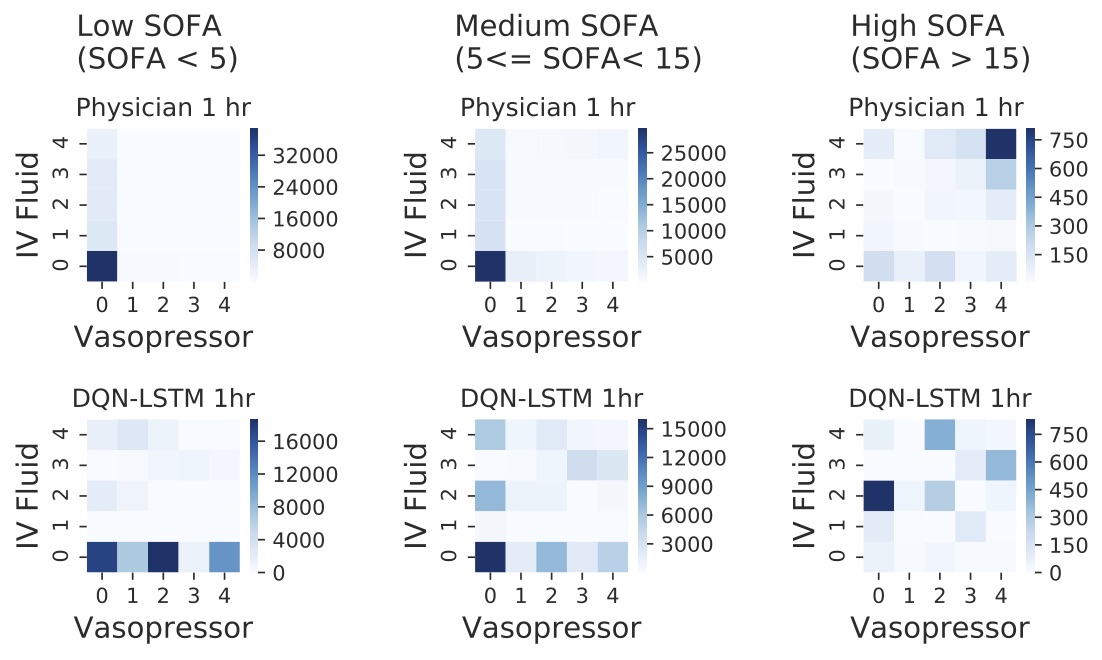}
    \captionsetup{font=small}
    \caption{Subgroup analysis: For patients with SOFA $<$ 5, both baseline and GRU implementation are 5-7 times more likely than physicians to give patients vasopressors.}
    \label{Subgroup_analysis}
\end{wrapfigure}

\textsl{\textit{\textbf{Subgroup Analysis: Grouping by SOFA score finds DQN agents are underaggressive in high risk patients and overaggressive in low risk patients}}}

\vspace{-0.5em}
We further perform an analysis in subgroups defined by severity of sepsis as indicated by Sequential Organ Failure Assessment (SOFA score\cite{Serial_Evaluation_of_the_SOFA_Score}). The SOFA score is a commonly used tool to stratify and compare patients in clinical practice, with higher scores indicating worse condition. When the assessment is greater than 15, mortality is increased up to 80\%.\cite{Serial_Evaluation_of_the_SOFA_Score} In Figure \ref{Subgroup_analysis}, the Dueling DDQN agents are significantly more aggressive than physicians in treating lower risk patients. As was also observed in Raghu \textit{et al}\cite{DRL_for_sepsis}, in high risk patients the reverse is true. Physicians commonly give maximum doses of both vasopressors and fluids, while the DQN agents rarely do. This suggests that the Dueling DDQN models may not be correctly accounting for patient severity or adjusting for confounding by indication.


\section*{Conclusion and Discussion}\label{conclusion}
State-of-the-art deep reinforcement learning approaches are largely untested in rich and dynamic healthcare settings.
We presented a sensitivity analysis exploring how a Dueling DDQN agent would react to alternative specifications, including: 
1) approaches to adjusting for treatment history;  2) discretized time bin durations; 3) recurrent neural network state representations; 4) reward specifications; and 5) random seeds. We have shown that choices between equally a priori justifiable implementation settings can have large clinically significant impacts on learned DQN policies. Given this lack of robustness, results from individual implementations should be received skeptically.  



The one area where our results do seem to point toward some clear guidance concerns the inclusion of treatment history in the state. 
Exclusion of treatment history from the state is only warranted under the implausible Markov assumption that past treatments only influence future outcomes through measured intermediate variables. If this assumption fails and there are cumulative dangers from too much treatment (e.g. pulmonary edema from fluid overload and cardiovascular side effects from vasopressors in our application), then past treatment will affect both the response (treatment is less likely to be beneficial given excessive past treatment) and the current treatment decision (treatment is less likely to be administered given extensive past treatment, and outcomes are likely to be worse given extensive past treatment). Thus, omitting treatment history would make excessive treatment appear more beneficial than it actually is. Indeed, the behavior we observed in our Dueling DDQN agents was consistent with the behavior that would be predicted by theory. Agents trained without treatment history included in their states recommended vasopressors or fluids at every timestep, an obviously harmful strategy. Yet all three prior DQN implementations in sepsis omitted treatment history from state definitions.

While we cannot provide definitive statistical or physiological explanations for most of the DQN outputs observed in our sensitivity analysis, here we discuss possible sources of DQN instabilities. One theme appeared to be unreasonable policies stemming from extrapolation beyond treatment decisions observed in the data under current practice. 
 For example, in our SOFA score subgroup analysis we saw the DQN agents recommending clearly harmful actions rarely seen in the data, i.e. failing to give high doses to the highest risk patients and frequently giving high doses to low risk patients. Also, the fact that different initializations found solutions with similar estimated Q-values but qualitatively different action distributions (also observed in Arjumand \textit{et al}\cite{Using_Maximum_Mean_Discrepancy_to_Find_a_Set_of_Diverse_Policies}) suggests that the problem is not sufficiently constrained.
We need better ways to incorporate knowledge of what features are important and what actions are reasonable to constrain learned policies\cite{off_policy_RL_without_Exploration}.

There is a long road from the current state of DRL healthcare applications to clinically actionable insights or treatment recommendations. Currently, AI researchers apply DRL algorithms to clinical problems and claim that the policies they learn would greatly improve health outcomes compared to current practice\cite{RLsepsis,DRL_for_sepsis}. In this work, we demonstrate that had these researchers made slightly different (but a priori reasonable) decisions, they would have obtained very different policies that also appeared superior to current practice. Beginning to map this sensitivity is a small but important step along the road to clinically actionable DRL policies. We hope our observations will lead to future work on characterizing and (more importantly) obtaining the type of robustness required to justify empirical testing of a DRL policy via a clinical trial.

We close with some speculative suggestions for that future work.  
First, a wide range of analysis settings can be compared in extensive and physiologically faithful simulation experiments where the ground truth value of resulting learned policies would be available. This could shed light on certain operating characteristics and best practices. For example, alternative approaches to preventing models from extrapolating too far from current practice as suggested above could be evaluated in this framework.

Further, when policies are sensitive to algorithmic choices, one could search for areas of broad agreement across policies recommended under a range of settings. These areas of agreement would be policy fragments, i.e. recommendations only applying to specific contexts. These strategy fragments could then be rigorously assessed by subject matter experts for plausibility and their effects could be estimated by epidemiologists using more stable techniques for treatment effect estimation. Seeking approaches to make DRL robust and human-verifiable will help us properly leverage information in health records to improve care.


\makeatletter
\renewcommand{\@biblabel}[1]{\hfill #1.}
\makeatother

\bibliographystyle{unsrt}

\begin{thebibliography}{1}
\setlength\itemsep{-0.1em}



\bibitem{SARSA}
M. Komorowski, A Gordon, L.A. Celi, A. Faisal, “A Markov decision process to suggest optimal treatment of severe infections in intensive care,” in Neural Information Processing Systems Workshop on Machine Learning for Health, 2016.

\bibitem{AI_clinician}
M. Komorowski, L.A. Celi, O. Badawi et al. The Artificial Intelligence Clinician learns optimal treatment strategies for sepsis in intensive care. Nat Med 24, 1716–1720 (2018). 

\bibitem{DRL_for_sepsis}
Aniruddh Raghu, M. Komorowski, L.A. Celi, Peter Szolovits, Marzyeh Ghassemi. Deep Reinforcement Learning for Sepsis Treatment. Machine Learning For Health at the conference on Neural Information Processing Systems, 2017;arXiv:1711.09602

\bibitem{continouseState}
Aniruddh Raghu , M. Komorowski, Imran Ahmed, Leo Celi, Peter Szolovits,   Marzyeh Ghassemi. Continuous State-Space Models for Optimal Sepsis Treatment - a Deep Reinforcement Learning Approach 2017;arXiv:1705.08422

\bibitem{RLsepsis}
Peng X, Ding Y, Wihl D, et al. Improving Sepsis Treatment Strategies by Combining Deep and Kernel-Based Reinforcement Learning. AMIA Annu Symp Proc. 2018;2018:887–896. 




\bibitem{RLonVentilation}
Niranjani Prasad, Li-Fang Cheng, Corey Chivers, Michael Draugelis, and Barbara E Engelhardt. A reinforcement learning approach to weaning of mechanical ventilation in intensive care units. 2017;arXiv:1704.06300 

\bibitem{HIV_treatment_selection}
Parbhoo S, Bogojeska J, Zazzi M, Roth V, Doshi-Velez F. Combining Kernel and Model Based Learning for HIV Therapy Selection. AMIA Jt Summits Transl Sci Proc. 2017;2017:239–248.

\bibitem{Atari}
Volodymyr Mnih, Koray Kavukcuoglu, David Silver et al; DeepMind Technologies. Playing Atari with Deep Reinforcement Learning; NIPS Deep Learning Workshop 2013; arXiv:1312.5602

\bibitem{DQN}
Volodymyr Mnih, Koray Kavukcuoglu, David Silver et al. Human-level control through deep reinforcement learning. Nature 518, 529–533 (2015).

\bibitem{double_qlearning}
Hado V. Hasselt. Double Q-learning. Advances in Neural Information Processing Systems 23; (2010);Pages 2613–2621

\bibitem{Dueling_DQN}
Ziyu Wang, Tom Schaul, Matteo Hessel, Hado van Hasselt, Marc Lanctot, Nando de Freitas. Dueling Network Architectures for Deep Reinforcement Learning; arXiv:1511.06581

\bibitem{survivingSepsisCampaign}
Andrew Rhodes, Laura E. Evans, Waleed Alhazzani et al. Surviving Sepsis Campaign: International Guidelines for Management of Sepsis and Septic Shock: 2016. Intensive Care Med 43, 304–377 (2017).

\bibitem{Inotropes_and_Vasopressors}
Overgaard Christopher B., Džavík Vladimír. Inotropes and Vasopressors. 2008; Circulation. 2008;118:1047–1056

\bibitem{evaluatingRLonHealth}
Omer Gottesman1, Fredrik Johansson, Joshua Meier1, Jack Dent1, et al.
Evaluating Reinforcement Learning Algorithms in Observational Health Settings 2018arXiv:1805.12298 

\bibitem{Brief_surcey_DRL}
K. Arulkumaran, M. P. Deisenroth, M. Brundage and A. A. Bharath. Deep Reinforcement Learning: A Brief Survey. IEEE Signal Processing Magazine, vol. 34, no. 6, pp. 26-38, Nov. 2017.

\bibitem{Qlearning}
Chris Watkins and Peter Dayan. Technical Note Q-Learning. Machine Learning 1992; 8, 279-292. 

\bibitem{prior_experience_relay}
Tom Schaul, John Quan, Ioannis Antonoglou, David Silver. Prioritized experience replay. arXiv preprint arXiv:1511.05952, 2015



\bibitem{MIMIC}
Alistair E.W. Johnson, Tom J. Pollard, Lu Shen et al. MIMIC-III, a freely accessible critical care database. Sci Data 3, 160035 (2016).
 
\bibitem{Sepsis-3}
M. Singer, C. S. Deutschmann, C. W. Seymour et al. The Third
International Consensus Definitions for Sepsis and Septic Shock
(Sepsis-3). JAMA. 2016;315(8):801-810. doi:10.1001/jama.2016.0287
 


\bibitem{time_varying_instructional_treatments}
Guanglei Hong and Stephen W. Raudenbush Causal Inference for Time-Varying Instructional Treatments.Journal of Educational and Behavioral Statistics, 33(3), pp. 333–362. doi: 10.3102/1076998607307355. 2008


\bibitem{LSTM}
Sepp Hochreiter Long short-term memory. Neural computation, 1997. 9(8): p. 1735-1780.



\bibitem{Robust_estimation}
James M Robins. Robust estimation in sequentially ignorable missing data and causal inference models. Biometrics 61, 962–972 December 2005; DOI: 10.1111/j.1541-0420.2005.00377.x



\bibitem{recurrent_auto_encoder_for_multidimentional_time_series_representation}
Zhengping Che, Sanjay Purushotham1, Kyunghyun Cho et al. Recurrent Neural Networks for Multivariate Time Series with Missing Values. Sci Rep 8, 6085 (2018). 


\bibitem{Empirical_Evaluation_of_Gated_Recurrent_Neural_Networks_on_Sequence_Modeling}
Junyoung Chung, Caglar Gulcehre, KyungHyun Cho, Yoshua Bengio. Empirical Evaluation of Gated Recurrent Neural Networks on Sequence Modeling; NIPS Deep Learning and Representation Learning Workshop 2014; arXiv:1412.3555



\bibitem{Reinforcement Learning with Long Short-Term Memory}
Bram Baker. Conference on Neural Information Processing Systems 2002 Dept.of Psychology,LeidenUniversity/  IDSIA 2002


\bibitem{GRU}
Kyunghyun Cho, Bart van Merriënboer, Dzmitry Bahdanau, Yoshua Bengio; On the Properties of Neural Machine Translation: Encoder:Decoder Approaches; Eighth Workshop on Syntax, Semantics and Structure in Statistical Translation 2014;DOI:10.3115/v1/W14-4012

\bibitem{Deep_Reinforcement_Learning_that_Matters}
Henderson Peter, Islam Riashat,  Bachman Philip, Pineau Joelle ,Precup Doina, Meger David. Deep Reinforcement Learning that Matters.AAAI Conference On Artificial Intelligence (AAAI) 2018; arXiv preprint arXiv:1709.06560





\bibitem{DRLPolicyEvaluation}
N. Jiang, Lihong Li
Microsoft Doubly Robust Off-policy Evaluation for Reinforcement Learning; arXiv:1511.03722 

\bibitem{Serial_Evaluation_of_the_SOFA_Score}
Flavio Lopes Ferreira, MD; Daliana Peres Bota, MD; Serial Evaluation of the SOFA Score to Predict Outcome in Critically Ill Patients. Annette Bross, MD; et al. 2001 The Journal of the American Medical Association 

\bibitem{Using_Maximum_Mean_Discrepancy_to_Find_a_Set_of_Diverse_Policies}
Masood, Muhammad and Doshi-Velez, Finale. (2019). Diversity-Inducing Policy Gradient: Using Maximum Mean Discrepancy to Find a Set of Diverse Policies. 5923-5929. 10.24963/ijcai.2019/821. 

\bibitem{off_policy_RL_without_Exploration}
Fujimoto Scott, Meger David, and Precup Doina. (2018). Off-Policy Deep Reinforcement Learning without Exploration. arXiv:1812.02900. 







\end{thebibliography}
{\footnotesize
\centering
}

\end{document}


\section{Appendix}
\textbf{Table 5 \& 6} demonstrate the action distribution of DQN-LSTM (baseline) and DQN-GRU under 1-hour window of vasopressor and fluid resuscitation. We calculate the 95\% CI of the mean difference by performing bootstrap resampling.
\begin{table}[H]
\caption{Action distribution of Vasopressor. }

\centering
\begin{tabular}{llllllll}
\toprule
{} & Physician & \multicolumn{3}{l}{DQN-LSTM }  &   \multicolumn{3}{l}{DQN-GRU}  \\    
\cmidrule(r){2-8} 
{} & No.\% & No.\%  & \multicolumn{2}{l}{95\% CI}& No.\%   & \multicolumn{2}{l}{95\% CI}  \\ \midrule
No action &   83.70  &  50.02 & 48.66 & 51.48 & 24.94 &  23.80 & 26.11 \\  
1st &   4.46  & 46.81 & 45.40 & 48.01 & 64.37 & 62.87 & 65.52 \\   
2nd &   4.23   & 0.0028 & 0.0016 &  0.004 &0.0039 & 0.0026 & 0.0058 \\   
3rd &   3.43 & 0.0240 & 0.020 & 0.0283 & 0.099 & 0.092 & 0.107 \\  
4th &   4.15 & 0.0051 &   0.0035 &0.0073 & 0.0034 & 0.0024 & 0.0049  \\ \bottomrule
\end{tabular}

\end{table}

\begin{table}[H]
\caption{Action distribution of IV fluid}
\centering
\begin{tabular}{llllllll}
\toprule
{} & Physician & \multicolumn{3}{l}{DQN-LSTM }  &   \multicolumn{3}{l}{DQN-GRU}  \\    
\cmidrule(r){2-8} 
{} & No.\% & No.\%  & \multicolumn{2}{l}{95\% CI}& No.\%   & \multicolumn{2}{l}{95\% CI}  \\ \midrule
No action &   60.12  &  73.20 & 72.04 & 74.42 & 75.14 &  74.08 & 76.22 \\  
1st &   10.06  & 25.28 & 24.15 & 26.49 & 0.047 & 0.056 & 0.064 \\   
2nd &   9.75  & 0.0011 & 0.0005 &  0.0019 &0.015 & 0.013 & 0.019 \\   
3rd &   10.20 & 0.0048 & 0.0031 & 0.0068 & 0.136 & 0.146 & 0.154 \\    
4th &   9.84  & 0.0088 &   0.0066 & 0.0113 & 0.027 & 0.030 & 0.033  \\ 
\bottomrule
\end{tabular}
\end{table}

\begin{table}[H]
\centering
\caption{Relative Risk of Treatment of the baseline and DQN-GRU  }
\begin{tabular}{lllll} 
\toprule
{}& \multicolumn{2}{l}{IV fluid} &\multicolumn{2}{l}{Vaospressor } \\
\cmidrule{2-3}\cmidrule{4-5}
  {} & Relative Risk & 95\% CI  & Relative Risk & 95\% CI \\   
  \midrule
	No action & 1.027 &  1.001 - 1.053 & 0.496 & 0.472 - 0.520 \\
	1st & 0.219 & 0.190 - 0.253 & 1.374 & 1.335 - 1.418 \\
	2nd & 12.63 & 7.360 - 36.75 & 1.387 & 0.792 - 2.807 \\
	3th & 27.73 & 18.91 - 45.42   &4.191 &3.565 - 4.992 \\
	4th & 3.367 & 2.595 - 4.545 & 0.678 &0.472 - 1.077 \\
\bottomrule
\end{tabular}

\end{table}

\textbf{Table 7 \& 8} demonstrate the action distribution among 4-hour and 1-hour time window. The difference is calculated by deduction of action distribution, bootstrapped result, of each quantile in different time window, $Action_{4hr} - Action_{1hr}$. 

\begin{table}[!h]
\centering
    \caption{ Action distribution of IV fluid - 4-hour VS 1-hour}
    \begin{tabular}{llll} 
    \toprule
    {}&   Physician(\%) &  LSTM & GRU  \\   \midrule
    No action & -9.51  & -27.56  &  -46.86\%  \\
    1st &  +2.17 &+14.91 &  +6.86  \\
    2nd &  +2.48  &  +11.17  & +5.62  \\
    3rd & +2.23  & +11.91  & +26.81  \\
    4th &  +2.74  & +19.42 & +7.70  \\
    \bottomrule
    \end{tabular}
\end{table}

\begin{table}[!h]
    \centering
    \caption{ Distribution of Vasopressor 4-hour VS 1-hour}
 	\begin{tabular}[H]{llll}
    \toprule    
{} &  Physician(\%)  &  LSTM (\%) &  GRU (\%)  \\  \midrule
    No action &   -10.03 &  -15.888 & +21.19  \\  
    1st &   +3.05 & -35.687 & -58.05  \\  
    2nd &    +1.81 &  +19.57 &  +9.13  \\   
    3rd &    +2.68 & +13.89 & +21.22 \\ 
    4th &    +2.49 &  +18.9 &   +6.54  \\   \bottomrule
\end{tabular}
\end{table}
\begin{table}[H]
        \centering
        \caption{Initiation rate of Vasopressor }
        \begin{tabular}{cccc } 
        \toprule   
	   Hour-bin &  Physician(\%) & Vasopressor(\%) & IV Fluid(\%) 
	   \\ 
	   \cmidrule(r){1-4}
        4-hour &  1.537 & 3.77 & 3.246   \\ 
        1-hour  &  0.630 & 2.50  & 2.12   \\	
	    \bottomrule
        \end{tabular}
        
        \end{table}
        
\begin{table}[H]
\centering
\caption{Distribution of Vasopressor 4-hour VS 1-hour }
\begin{tabular}{lllll} 
\toprule
{}& \multicolumn{2}{l}{Vasopressor} &\multicolumn{2}{l}{IV Fluid } \\
\cmidrule{2-3}\cmidrule{4-5}
  {} & IR & 95\% CI  & IR & 95\% CI \\   
  \midrule
	4-hour  & 3.77  & 3.21 - 4.33 & 3.24 & 2.91 - 3.57 \\
	1-hour & 2.50 & 1.92 - 3.08 & 2.12 & 1.87 - 2.37 \\
\bottomrule
\end{tabular} 
\end{table}